\RequirePackage{fix-cm}
\documentclass[twocolumn]{svjour3}          % twocolumn
\smartqed  % flush right qed marks, e.g. at end of proof

\usepackage{booktabs}
\usepackage{multirow}
\usepackage{cite}
\usepackage[pdftex]{graphicx}
\usepackage{ragged2e}
\usepackage[tight,footnotesize]{subfigure}
\usepackage{rotating}
\usepackage{graphicx}
\usepackage{amsmath,amssymb} % define this before the line numbering.
\usepackage{array}
\usepackage{multirow}
\usepackage{colortbl}
\usepackage{amsfonts}
\usepackage{pifont}
\usepackage{xspace}
\usepackage{etoolbox}
\usepackage{overpic}
\usepackage{color}
\usepackage{microtype}

\newcommand{\orcid}[1]{\href{https://orcid.org/#1}{\includegraphics[width=10pt]{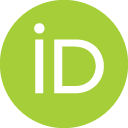}}}

\usepackage{hyperref}
\hypersetup{breaklinks=true,citecolor=blue, colorlinks}

\graphicspath{{./Imgs/}}
\DeclareGraphicsExtensions{.pdf,.jpg,.png}

\usepackage{silence}
\journalname{Research Article}

\begin{document}

\title{Multimodal Sentiment Analysis: A Survey}

\titlerunning{Multimodal Sentiment Analysis: A Survey}        % For running head

\author{Songning Lai \orcid{0009-0007-3132-9414 }        \and
  Xifeng Hu \orcid{0000-0001-7825-0710} \and
  Haoxuan Xu \orcid{0009-0006-6435-8900} \and 
  Zhaoxia Ren \and
  Zhi Liu \orcid{0000-0002-7640-5982}
}

% \author{first \orcid{0000-0000-0000-0000 }        \and
%   second \orcid{0000-0000-0000-0000} \and 
%   third \orcid{0000-0000-0000-0000}
% }

% \authorrunning{Songning Lai \orcid{0009-0007-3132-9414 }        \and
%   Haoxuan Xu \orcid{0000-0000-0000-0000} \and 
%   Zhi Liu \orcid{0000-0002-7640-5982}} % if too long for running head

\institute{
Songning Lai (\url{202000120172@mail.sdu.edu.cn}), Xifeng Hu (\url{201942544@mail.sdu.edu.cn}), Haoxuan Xu (\url{202020120237@mail.sdu.edu.cn}) and Zhi Liu (\url{liuzhi@sdu.edu.cn}) are with the School of Information Science and Engineering, Shandong University, Qingdao, China.\\
Zhaoxia Ren (\url{renzx@sdu.edu.cn}) is with Assets and Laboratory Management Department, Shandong University,  Qingdao, China.\\
% (Email: xxx@xxx.xx.xx, xxx@xxx.xx.xx). \\
% Third Author is with institution, (department), city, (state), country. 
% (Email: xxx@xxx.xx.xx). \\
Corresponding author: Zhi Liu and Zhaoxia Ren.
(\url{liuzhi@sdu.edu.cn},\url{renzx@sdu.edu.cn}). 
}

% \institute{
% first author are with the School of Information Science and Engineering, Shandong University, Qingdao 266237, China.\\
% (Email: xxx@xxx.xx.xx, xxx@xxx.xx.xx). \\
% Third Author is with institution, (department), city, (state), country. 
% (Email: xxx@xxx.xx.xx). \\
% Corresponding author:(Email: xxx@xxx.xx.xx). 
% }

 \date{Received: date / Accepted: date}
% % The correct dates will be entered by the editor

\maketitle

\begin{abstract}
Multimodal sentiment analysis has become an important research area in the field of artificial intelligence. With the latest advances in deep learning, this technology has reached new heights. It has great potential for both application and research, making it a popular research topic. This review provides an overview of the definition, background, and development of multimodal sentiment analysis. It also covers recent datasets and advanced models, emphasizing the challenges and future prospects of this technology. Finally, it looks ahead to future research directions. It should be noted that this review provides constructive suggestions for promising research directions and building better performing multimodal sentiment analysis models, which can help researchers in this field.

% Please provide 4 to 6 keywords which can be used for indexing purposes.

\end{abstract}

\section{Introduction}
\label{sec:intr}
Emotion is a subjective reaction of an organism to external stimuli \cite{plutchik2001nature,deonna2012emotions}. Humans possess a powerful capacity for sentiment analysis, and researchers are currently exploring ways to make this ability available to artificial agents \cite{hutto2014vader}.

Sentiment analysis involves analyzing sentiment polarity through available information \cite{kim2004determining,cambria2013new}. With the rapid development of fields such as artificial intelligence, computer vision, and natural language processing, it is becoming increasingly possible for artificial agents to implement sentiment analysis. Sentiment analysis is an interdisciplinary research area that includes computer science, psychology, social science, and other fields \cite{parvaiz2023vision,zhang2023toward,liu2023pre}.

Scientists have been working to empower AI agents with sentiment analysis capabilities for decades. This is a key component of human-like AI, making AI more like a human.

Sentiment analysis has significant research value \cite{chan2023state,wankhade2022survey,li2022word,yadav2020sentiment}. With the explosive growth of Internet data, vendors can use evaluative data such as reviews and review videos to improve their products. Sentiment analysis also has countless research values, such as lie detection, interrogation, and entertainment. The following sections will elaborate on the application and research value of sentiment analysis.

In the past, sentiment analysis has mostly focused on a single modality (visual modality, speech modality, or text modality) \cite{chandrasekaran2021multimodal}. Text-based sentiment analysis \cite{kratzwald2018decision,strapparava2007semeval,li2018generative} has gone a long way in NLP. Vision-based sentiment analysis pays more attention to human facial expressions \cite{DaiRong} and movement postures. Speech-based sentiment analysis mainly extracts features such as pitch, timbre, and temperament in speech for sentiment analysis \cite{ren2014acoustics}. With the development of deep learning, these three modalities have gained some foothold in sentiment analysis.

\begin{figure*}[t]
  \includegraphics[width=1.0\linewidth]{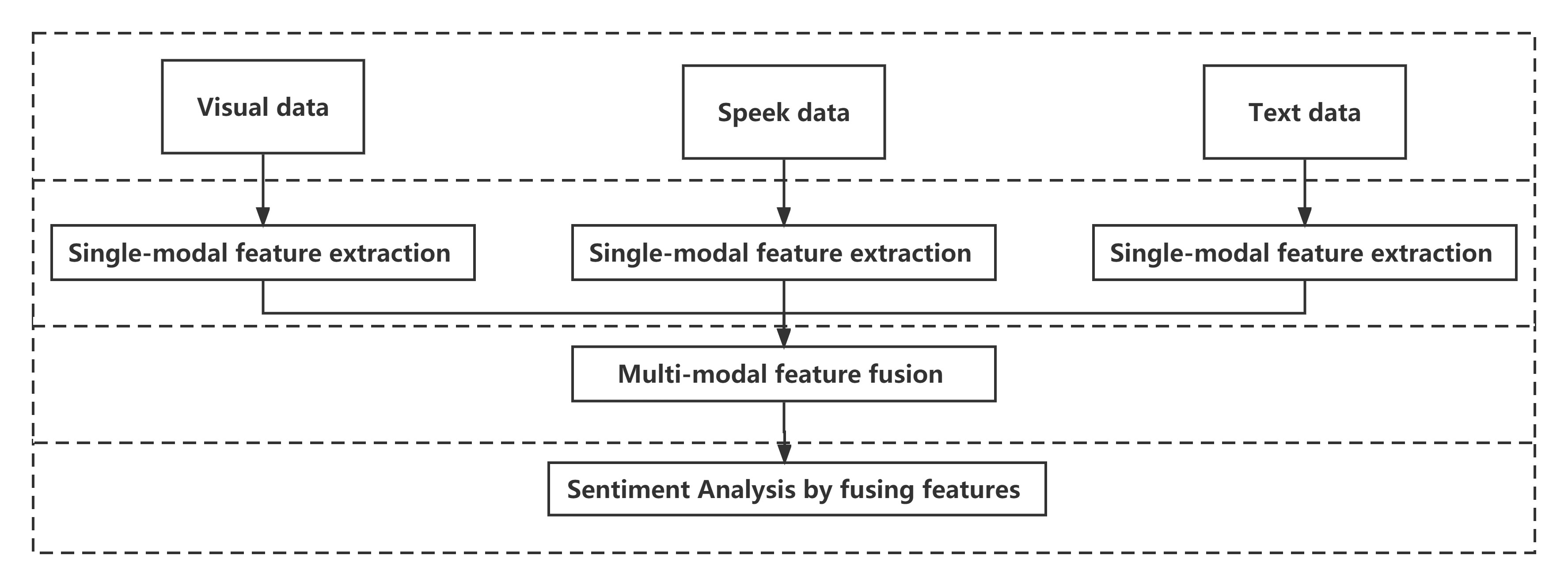}
  \caption{Figure shows the model architecture for the more classical multi-modal sentiment analysis. The overall architecture consists of three parts: one part for feature extraction of individual modalities, one part for fusion of features of each modality, and one part for sentiment analysis of the fused features. These three parts are very important, and researchers have begun to optimize these three parts one by one}
  \label{fig:2}       % Give a unique label
\end{figure*}

However, using a single modality for sentiment analysis has limitations \cite{jiming2021summary,huang2019image,kumar2019sentiment,gandhi2021multimodal}. The emotional information contained in a single modality is limited and incomplete. Combining information from multiple modalities can provide deeper emotional polarity. Analyzing only one modality results in limited results and makes it difficult to accurately analyze the emotion of an action.

Researchers have gradually realized the need for multi-modal sentiment analysis, and many multi-modal sentiment analysis models have emerged to accomplish this task. Text features dominate and play a key role in the analysis of deep emotions\cite{rupapara2021impact}. Visual modality extraction of expression and pose features can effectively aid text sentiment analysis and judgment\cite{li2022hybrid}. On the one hand, speech modality can extract text features, and on the other hand, speech tone can be recognized to reveal the state of text at each time point\cite{favaro2023multi}. Figure\ref{fig:2} shows the model architecture for the more classical multi-modal sentiment analysis. The overall architecture consists of three parts: one part for feature extraction of individual modalities, one part for fusion of features of each modality, and one part for sentiment analysis of the fused features. These three parts are very important, and researchers have begun to optimize these three parts one by one \cite{poria2017review}. 

In this review, we provide a comprehensive overview of the field of multimodal sentiment analysis. The review includes a summary and brief introduction of datasets, which can help researchers select appropriate datasets. We compare and analyze models that have significant research significance in multimodal sentiment analysis and provide suggestions for model construction. We elaborate on three types of modal fusion methods and explain the advantages and disadvantages of different modal fusion methods. Finally, we look ahead to the challenges and future development directions of multimodal sentiment analysis, providing several promising research directions. Compared to other reviews in the same field, our focus is on providing constructive suggestions for promising research directions and building better performing multimodal sentiment analysis models. We emphasize the challenges and future prospects of these technologies.

\section{Multimodal Sentiment Analysis Datasets}

With the growth of the Internet, an era of data explosion has been created\cite{munirathinam2020industry,ortiz2023rise,haseeb2019does}.  Numerous researchers have widely collected these data from the Internet (videos, reviews, etc.) and built sentiment datasets according to their own needs. Tab\ref{tab:addlabel} summarizes the commonly used multimodal datasets. The first column indicates the name of the data set. The second column is the year in which the sentiment data was released. The third column is the category of modalities included in the sentiment dataset. The fourth column is the platform from which the data set came. The fifth column is the language used by the dataset.The sixth column is the amount of data contained in the dataset.Each dataset has its own characteristics. This section lists well-known datasets in the community, aiming to help researchers sort out the characteristics of each dataset and make it easier for them to choose datasets.

\begin{table*}[htbp]
  \centering
    
    \begin{tabular}{llllllll}
    \toprule
    \multicolumn{2}{l}{\textbf{Name}} & \textbf{Year} & \multicolumn{2}{l}{\textbf{Modalities}} & \textbf{Source} & \textbf{Language} & \textbf{Number} \\
    \midrule
    \multicolumn{2}{l}{IEMOCAP} & 2008  & \multicolumn{2}{l}{A+V+T} & N/A   & English & 10039 \\
    \midrule
    \multicolumn{2}{l}{\multirow{3}[2]{*}{DEAP}} & \multirow{3}[2]{*}{2011} & \multicolumn{2}{l}{A+V+T} & \multirow{3}[2]{*}{N/A} & \multirow{3}[2]{*}{English} & \multirow{3}[2]{*}{10039} \\
    \multicolumn{2}{l}{} &       & \multicolumn{2}{l}{A+V+T} &       &       &  \\
    \multicolumn{2}{l}{} &       & \multicolumn{2}{l}{A+V+T} &       &       &  \\
    \midrule
    \multicolumn{2}{l}{CMU-MOSI} & 2016  & \multicolumn{2}{l}{A+V+T} & YouTube & English & 2199 \\
    \midrule
    \multicolumn{2}{l}{CMU-MOSEI} & 2018  & \multicolumn{2}{l}{A+V+T} & YouTube & English & 23453 \\
    \midrule
    \multicolumn{2}{l}{MELD} & 2019  & \multicolumn{2}{l}{A+V+T} & The Friends & English & 13000 \\
    \midrule
    \multicolumn{2}{l}{Multi-ZOL} & 2019  & \multicolumn{2}{l}{V+T} & ZOL.com & Chinese & 5288 \\
    \midrule
    \multicolumn{2}{l}{CH-SIMS} & 2020  & \multicolumn{2}{l}{A+V+T} & N/A   & Chinese & 2281 \\
    \midrule
    \multicolumn{2}{l}{\multirow{4}[2]{*}{CMU-MOSEAS}} & \multirow{4}[2]{*}{2021} & \multicolumn{2}{l}{\multirow{4}[2]{*}{A+V+T}} & \multirow{4}[2]{*}{YouTube} & Spanish & \multirow{4}[2]{*}{40000} \\
    \multicolumn{2}{l}{} &       & \multicolumn{2}{l}{} &       & Portuguese &  \\
    \multicolumn{2}{l}{} &       & \multicolumn{2}{l}{} &       & German &  \\
    \multicolumn{2}{l}{} &       & \multicolumn{2}{l}{} &       & French &  \\
    \midrule
    \multicolumn{2}{l}{FACTIFY} & 2022  & \multicolumn{2}{l}{V+T} & Twitter & English & 50000 \\
    \midrule
    \multicolumn{2}{l}{\multirow{2}[2]{*}{MEMOTION}} & \multirow{2}[2]{*}{2022} & \multicolumn{2}{l}{\multirow{2}[2]{*}{V+T}} & Reddit & \multirow{2}[2]{*}{English} & \multirow{2}[2]{*}{10000} \\
    \multicolumn{2}{l}{} &       & \multicolumn{2}{l}{} & Facebok &       &  \\
    \bottomrule
    \end{tabular}%

    \caption{This table contains the used multimodal datasets. The first column indicates the name of the data set. The second column is the year in which the sentiment data was released. The third column is the category of modalities included in the sentiment dataset. The fourth column is the platform from which the data set came. The fifth column is the language used by the dataset.The sixth column is the amount of data contained in the dataset.}
  \label{tab:addlabel}%
\end{table*}%

\subsection{IEMOCAP\cite{busso2008iemocap}}
IEMOCAP, a sentiment analysis dataset released by the Speech Analysis and Interpretation Laboratory in 2008, is a multi-modal dataset that comprises 1,039 conversational segments, with a total video length of 12 hours. Participants in the study engaged in five different scenarios, performing emotions as per a pre-set scenario. The dataset includes not only audio, video, and text information but also facial expression and posture information obtained through additional sensors. Data points are categorized into ten emotions: neutral, happy, sad, angry, surprised, scared, disgusted, frustrated, excited, and other. Overall, IEMOCAP provides a rich resource for researchers exploring sentiment analysis across multiple modalities.

\subsection{DEAP\cite{koelstra2011deap}}
DEAP is a dataset specifically designed for sentiment analysis that utilizes physiological signals as its source of data (Koelstra et al., 2011). The dataset examines EEG data from 32 subjects, with a 1:1 ratio of male and female participants. EEG signals were collected at 512Hz from different areas of the subjects' brains, including the frontal, parietal, occipital, and temporal lobes. To annotate the EEG signals, the subjects were asked to rate the corresponding videos in terms of three aspects: Valence, Arousal, and Dominance, on a scale of 1 to 9. This dataset provides valuable resources for researchers to explore sentiment analysis using physiological signals.

\subsection{CMU-MOSI\cite{zadeh2016mosi}}
CMU-MOSI dataset is comprised of 93 critical YouTube videos that cover a range of topics (Zadeh et al., 2016). These videos were carefully selected to ensure that they featured only one speaker who was facing the camera, allowing for clear capture of facial expressions. While there were no restrictions on camera model, distance, or speaker scene, all presentations and comments were made in English by 89 different speakers, including 41 women and 48 men. The 93 videos were divided into 2,199 subjective opinion segments and annotated with sentiment intensity ranging from strongly negative to strongly positive (-3 to 3). Overall, the CMU-MOSI dataset provides a valuable resource for researchers studying sentiment analysis.

\subsection{CMU-MOSEI\cite{zadeh2018multimodal}}
CMU-MOSEI is a popular dataset for sentiment analysis that comprises 3,228 YouTube videos (Zadeh et al., 2018). These videos are categorized into 23,453 segments and feature data from three different modalities: text, visual, and sound. With contributions from 1,000 speakers and coverage of 250 different topics, this dataset offers a diverse range of perspectives. All the videos are in English, and both sentiment and emotion annotations are available. The six emotion categories include happy, sad, angry, scared, disgusted, and surprised, while the sentiment intensity markers range from strongly negative to strongly positive (-3 to 3). Overall, CMU-MOSEI is an invaluable resource for researchers exploring sentiment analysis across multiple modalities.

\subsection{MELD\cite{poria2018meld}}
MELD is a comprehensive dataset that includes video clips from the popular television series Friends. The dataset comprises textual, audio, and video information that corresponds to the textual data. It contains 1400 videos, which are further divided into 13,000 individual segments. The dataset is annotated with seven categories of annotations: Anger, Disgust, Sadness, Joy, Neutral, Surprise, and Fear. Each segment has three sentiment annotations: positive, negative, and neutral.

\subsection{Multi-ZOL\cite{xu2019multi}}
Multi-ZOL is a dataset designed for bimodal sentiment classification of images and text. The dataset consists of reviews of mobile phones collected from ZOL.com. It contains 5288 sets of multimodal data points that cover various models of mobile phones from multiple brands. These data points are annotated with a sentiment intensity rating from 1 to 10 for six aspects.

\subsection{CH-SIMS\cite{yu2020ch}}
CH-SIMS is a unique dataset consisting of 60 open-sourced videos from the web that are split into 2281 video clips. The dataset focuses solely on Chinese (Mandarin) language and ensures that each segment contains only one character's face and voice. It covers a wide range of scenes and speaker ages and is individually labeled for each modality, making it a valuable resource for researchers. The dataset annotations contain sentiment intensity ratings ranging from negative to positive (-1 to 1) and also include annotations for other attributes such as age and gender.

\subsection{CMU-MOSEA\cite{zadeh2020cmu}}
CMU-MOSEA is a versatile dataset that includes multiple languages, such as Spanish, Portuguese, German, and French. The dataset comprises 40,000 sentence fragments, covering 250 different topics and 1645 speakers. The annotations are split into two categories: sentiment intensity and binary. Each sentence is annotated with sentiment strength in the interval [-3,3], and the binary includes whether the speaker expressed an opinion or made an objective statement. Emotions are divided into six categories for each sentence: happiness, sadness, fear, disgust, surprise, and annotation.

\subsection{FACTIFY\cite{mishra2022factify}}
FACTIFY is a fake news detection dataset that focuses on implementation validation. It includes data for both image and text modalities and contains 50,000 sets of data. Most of the data's claims refer to politics and government. The dataset is annotated into three categories: support, no evidence, and refutation. This dataset is a valuable resource for researchers interested in detecting and combating the spread of fake news.

\subsection{MEMOTION\cite{ramamoorthy2022memotion}}
MEMOTION is a meme-based dataset that includes popular memes related to politics, religion, and sports. The dataset comprises 10,000 data points and is divided into three sub-tasks: Sentiment Analysis, Emotion Classification, and Scale/Intensity of Emotion Classes. The annotators annotated each data point differently under the different subtasks. Subtask one annotates each data point into three categories (negative, neutral, and positive). Subtask two annotates each data point into four categories (humour, sarcasm, offense, motivation). Subtask three annotates each data point in the interval [0,4] to indicate the sentiment intensity. This dataset provides a unique opportunity for researchers to analyze the use of memes as a means of communication and expression in modern culture.

The aforementioned datasets all have certain limitations. We summarize the limitations of each dataset to help researchers make informed decisions when selecting datasets for experiments. For IEMOCAP, the limited number of actors in the dataset may lead to overfitting issues. Additionally, the emotion categories in the dataset may not be comprehensive enough to cover all emotion types. For DEAP, CMU-MOSI, and CMU-MOSEI, the emotion categories in the datasets may not be comprehensive enough to cover all emotion types. Furthermore, the video clips in the datasets may be too short to fully express emotions. For MELD, the audio samples in the dataset may be too short to fully express emotions. Additionally, facial expressions may be affected by environmental factors such as lighting. For Multi-ZOL, since the sample sources for social media comments are diverse, there may be some degree of noise. Furthermore, this dataset is only suitable for sentiment analysis tasks on social media comments. For CH-SIMS, this dataset is only suitable for student negative emotion recognition tasks. Additionally, due to the limited sample sources being restricted to student questionnaires, there may be a certain degree of subjective bias. For CMU-MOSEA, this dataset is only suitable for multimodal information and emotion recognition tasks in movie scenes. Additionally, due to the limited sample sources being restricted to movie scene annotations, there may be a certain degree of subjective bias. For FACTIFY, this dataset is only suitable for topic and sentiment recognition tasks in news articles. Due to the limited sample sources being restricted to news article annotations, there may be a certain degree of subjective bias.

\section{Multimodal fusion}
Multimodal data describe objects from different perspectives and are more informative than single-modal data. Data information from different modalities can be complementary to each other. In the task of multimodal sentiment analysis, it is a very important and challenging task to fuse the data features between different modalities, preserve the semantic integrity of the modalities, and achieve a good fusion between different modalities. According to the different modes of modal fusion, it can be summarized as feature-based multimodal fusion in the early stage, model-based multimodal fusion in the middle stage, and decision-based multimodal fusion in the late stage.

\subsection{Early feature-based approaches for multimodal fusion}

Early feature-based multimodal fusion methods perform shallow fusion after feature extraction in the early stage. The fusion of the features of different modalities at the shallow level of the model is equivalent to unifying the features of different single modalities into the same parameter space. Due to the differences in information between different modalities, features commonly contain a large amount of redundant information, and it is frequently necessary to use dimensionality reduction methods to remove the redundant information. Features after dimensionality reduction are fed into the model to complete feature extraction and prediction. Early feature fusion wants the model to consider input information from multiple modalities at the beginning of feature modeling. However, the method of unifying multiple different parameter spaces in the input layer may not achieve the desired effect due to the differences in the parameter spaces of different modes. This category of models can effectively handle multimodal emotion recognition tasks with high accuracy and robustness. However, these models require a large amount of training data to achieve good performance, and their structures are relatively complex, which requires a longer training time. The overall framework of early feature-based approaches for multimodal fusion is shown in Figure\ref{fig:3}. Some representative models are:

\begin{figure}[t]
  \includegraphics[width=1.0\linewidth]{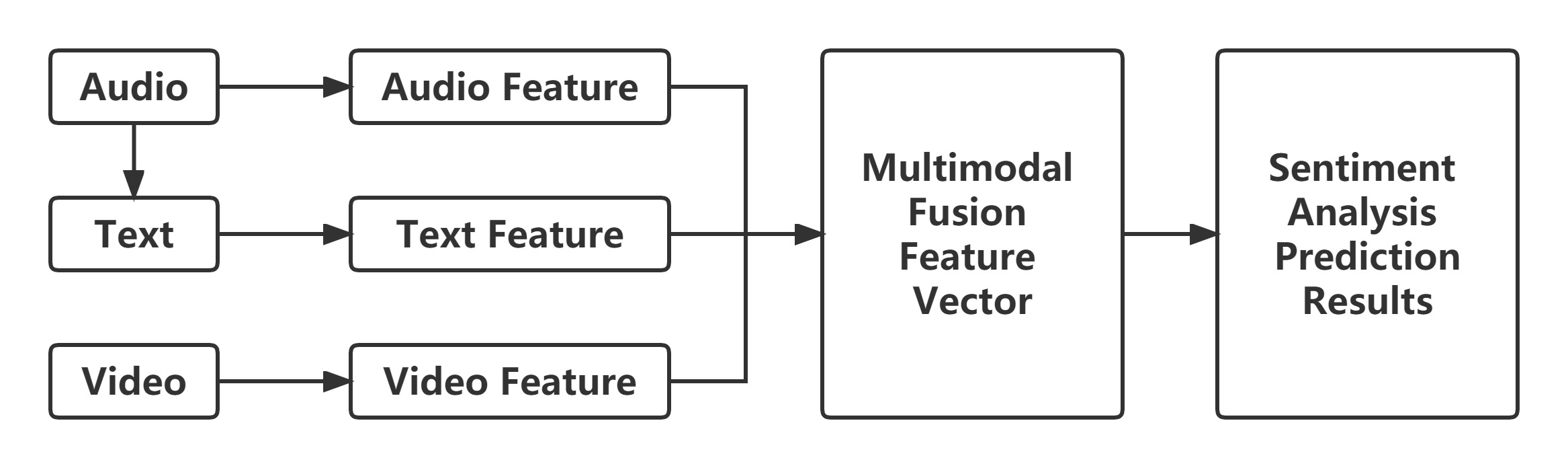}
  \caption{Figure shows the overall framework of early feature-based approaches for multimodal fusion. After extracting features, this model uses specific components to fuse the features of each modality.}
  \label{fig:3}       % Give a unique label
\end{figure}

\subsubsection{\textbf{THMM (Tri-modal Hidden Markov Model) \cite{morency2011towards}}} 
One approach to multimodal sequence modeling and analysis is to represent eigenvectors of multiple modalities as higher-order tensors and use tensor decomposition methods to extract hidden states and transition probabilities. This approach effectively exploits the correlation and complementarity between multimodal data, while avoiding the curse of dimensionality and overfitting. However, the disadvantage is that the order and rank of the tensors and the number of hidden states must be predetermined, which may affect the model's performance and efficiency.

\subsubsection{\textbf{RMFN (Recurrent Multistage Fusion Network) \cite{liang2018multimodal}}} Another approach is to use multiple recurrent neural network layers to gradually fuse features from different modalities, from local to global, from low-level to high-level, and finally obtain a comprehensive sentiment representation. This model uses an attention mechanism to adjust the location of features in semantic space for different modalities, allowing the same word to exhibit different emotions under different nonverbal behaviors.

\subsubsection{\textbf{RAVEN (Recurrent Attended Variation Embedding Network) \cite{wang2019words}}} A Hierarchical Fusion Network has been proposed for multimodal sentiment analysis, which includes a local fusion module and a global fusion module. Local cross-modal fusion is explored through a sliding window, which effectively reduces computational complexity.

\subsubsection{\textbf{HFFN(Hierarchical feature fusion network with local and global perspectives for multimodal affective computing) \cite{mai2019divide}}} Another approach is to use recurrent neural networks and adversarial learning to learn joint representations between different modalities, thereby improving the ability of single-modal representations and dealing with missing modalities or noise.

\subsubsection{\textbf{MCTN (Multimodal Cyclic Translation Network) \cite{pham2019found}}} A medium-term model-based multimodal fusion approach involves feeding multimodal data into the network, and the intermediate layers of the model perform feature fusion between the modalities. Model-based modality fusion methods can select the location of modality feature fusion to achieve intermediate interactions. Model-based fusion typically uses multiple kernel learning, neural networks, graph models, and alternative methods.

\subsection{Medium-term model-based multimodal fusion method}

\begin{figure}[t]
  \includegraphics[width=1.0\linewidth]{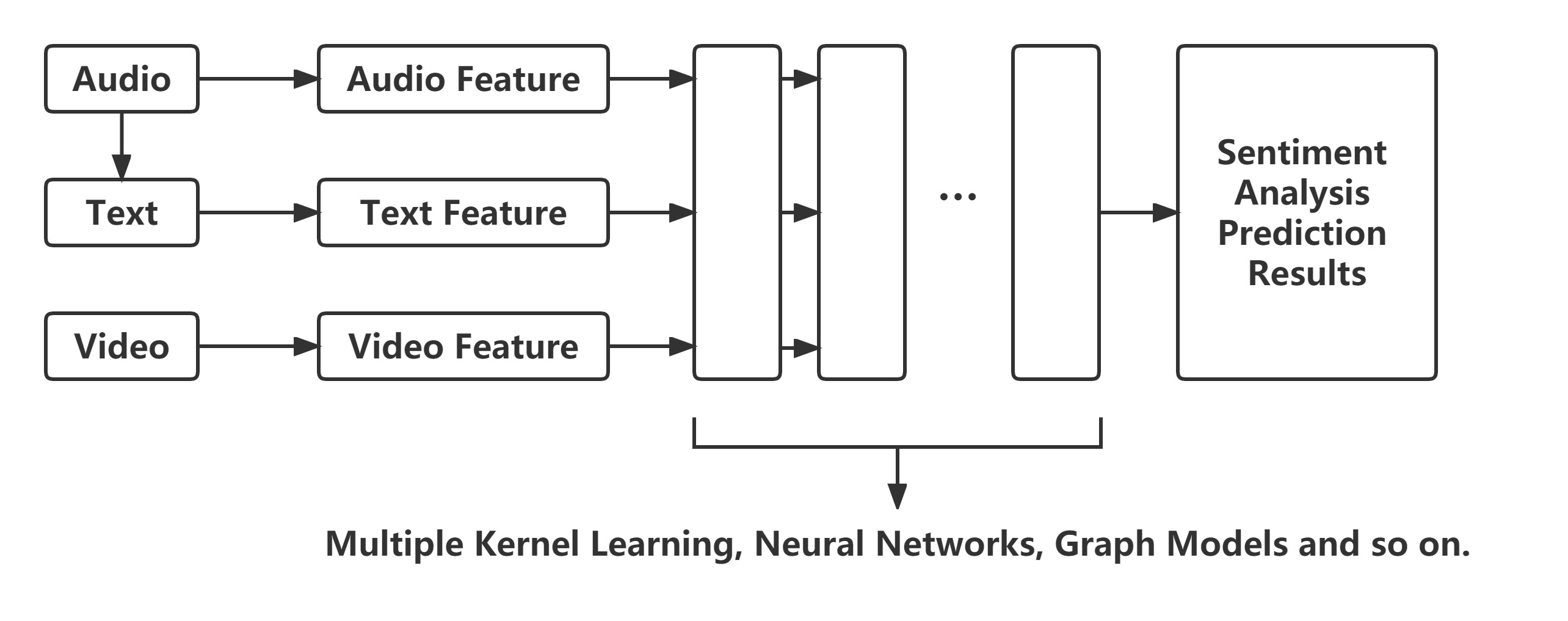}
  \caption{Figure shows the overall framework of medium-term model-based approaches for multimodal fusion. This class inputs the feature information of each modality into multiple kernel learning, neural networks, graph models, and alternative methods to complete the fusion of modalities. Most of the nodes of its modal fusion are variable.}
  \label{fig:4}       % Give a unique label
\end{figure}

A medium-term model-based multimodal fusion approach is to feed multimodal data into the network, and the intermediate layers of the model perform feature fusion between the modalities. Model-based modality fusion methods can select the location of modality feature fusion to achieve intermediate interactions. Model-based fusion typically uses multiple kernel learning, neural networks, graph models, and alternative methods. The overall framework of medium-term model-based multimodal fusion method is shown in Figure\ref{fig:4}.

\subsubsection{\textbf{MKL (Multiple Kernel Learning) \cite{poria2015deep}}} This model is a multiple kernel learning approach. It uses different kernel functions to represent different modal information and selects the optimal combination of kernel functions by optimizing an objective function to achieve the fusion of multi-modal information. The model has high flexibility and can adaptively select different kernel functions, thereby improving the robustness and accuracy of the model. However, the model structure is relatively simple and may not be able to handle complex emotional expressions.

\subsubsection{\textbf{BERT-like (Self Supervised Models to Improve Multimodal Speech Emotion Recognition) \cite{siriwardhana2020jointly}}} This model is a Transformer-based multi-modal sentiment analysis method that can leverage self-attention mechanism to achieve alignment and fusion between text and image. The model adopts a self-supervised learning method, which can effectively handle multimodal emotion recognition tasks with high accuracy and robustness. In addition, the model may be affected by the quality of data and annotations.

\subsection{Multimodal model based on decision fusion in the later stage}

\begin{figure}[t]
  \includegraphics[width=1.0\linewidth]{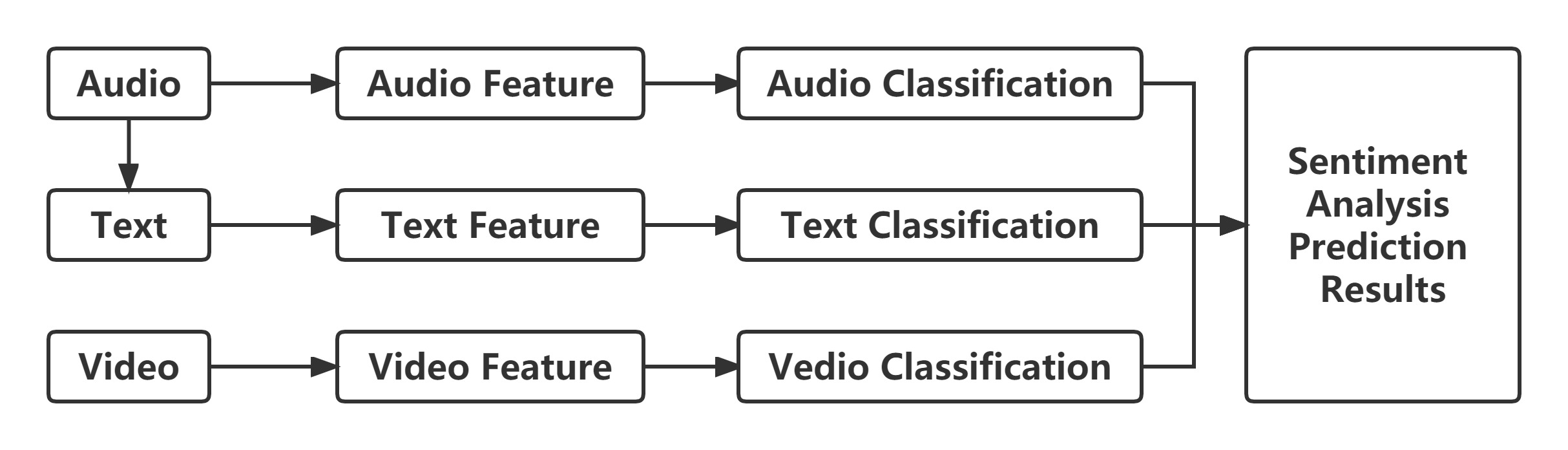}
  \caption{Figure shows the overall framework of multimodal model based on decision fusion in the later stagen method. These models train a separate classifier for each modality. Finally, the classifiers of each modality were integrated to complete the task of multimodal sentiment analysis.}
  \label{fig:5}       % Give a unique label
\end{figure}

A decision level fusion method is used to fuse information from different modalities. Decision-level fusion refers to training models separately on data from different modalities to incorporate outputs from different modalities into the final decision. Multimodal models based on decision fusion typically fuse modalities using methods such as averaging, majority voting, weighting, and learnable models. Such models are typically lightweight and flexible. When any modality is missing, the decision can be made by using the remaining modalities. The overall framework of multimodal model based on decision fusion in the later stagen method is shown in Figure\ref{fig:5}.

\subsubsection{\textbf{Deep Multimodal Fusion Architecture \cite{nojavanasghari2016deep}}} In this model, each modality has an independent classifier. The prediction results are output after averaging the confidence scores of each classifier. The model has a simple structure, is easy to implement, and can effectively handle multimodal emotion recognition tasks. However, the model cannot handle the interaction information between modalities and may suffer from information loss.

\subsubsection{\textbf{SAL-CNN (Select-Additive Learning CNN) \cite{wang2017select}}} This model is a multimodal sentiment analysis model based on CNN and attention mechanism. It uses an adaptive attention mechanism to fuse text and image features, a spatial attention mechanism to extract text-related regions in images, and finally a completely connected layer to classify the output. The model adopts an attention mechanism, which can effectively handle multimodal emotion recognition tasks with high accuracy and robustness. However, the model requires a large amount of training data to achieve good performance, and the model structure is relatively complex, requiring a longer training time.

\subsubsection{\textbf{TSAM (Temporally Selective Attention Model) \cite{yu2017temporally}}} The proposed model is a time-selective attention model, which assigns weights through an attention mechanism to help the model choose the time step, and finally sends it to a distinct SDL loss function model for sentiment analysis. SDL is a multi-modal sentiment analysis method based on Self-Distillation Learning, which can exploit the complementarity between different modalities to improve the generalization ability and robustness of the model. The model adopts a time-selective attention mechanism, which can effectively handle multimodal emotion recognition tasks with high accuracy and robustness.

\section{Latest Multimodal Sentiment Analysis Models}

In recent years, the field of multimodal sentiment analysis has evolved into a huge system and many practical and efficient models and architectures have emerged. It's hard to cover all the models here. In this chapter, we present some recent and cutting-edge multimodal sentiment analysis models. Most of these models were used as benchmark models by later researchers for experimental reference. These models are summarized in Tab\ref{tab:addlabe2}. The first column is the name of the model. The second column is the year in which the model was published. The third column is the dataset used by the model. The fourth column is the accuracy under this dataset.

\begin{table*}[htbp]
  \centering

    \begin{tabular}{lllllll}
    \toprule
    \textbf{Name} &       & \textbf{Year} & \multicolumn{2}{l}{\textbf{Dataset}} & \multicolumn{2}{l}{\textbf{Acc}} \\
    \midrule
    \multicolumn{2}{l}{MultiSentiNet-Att} & 2017  & \multicolumn{2}{l}{MVSA} & \multicolumn{2}{l}{68.86\%} \\
    \midrule
    \multicolumn{2}{l}{\multirow{2}[2]{*}{DFF-TMF}} & \multirow{2}[2]{*}{2019} & \multicolumn{2}{l}{CMU-MOSI} & \multicolumn{2}{l}{80.98\%} \\
    \multicolumn{2}{l}{} &       & \multicolumn{2}{l}{CMU-MOSEI} & \multicolumn{2}{l}{77.15\%} \\
    \midrule
    \multicolumn{2}{l}{\multirow{2}[2]{*}{AHRM}} & \multirow{2}[2]{*}{2020} & \multicolumn{2}{l}{Flickr} & \multicolumn{2}{l}{87.10\%} \\
    \multicolumn{2}{l}{} &       & \multicolumn{2}{l}{Getty Image} & \multicolumn{2}{l}{87.80\%} \\
    \midrule
    \multicolumn{2}{l}{SFNN} & 2020  & \multicolumn{2}{l}{Yelp} & \multicolumn{2}{l}{62.90\%} \\
    \midrule
    \multicolumn{2}{l}{MISA} & 2020  & \multicolumn{2}{l}{MOSI} & \multicolumn{2}{l}{83.40\%} \\
    \midrule
    \multicolumn{2}{l}{\multirow{2}[2]{*}{MAG-BERT}} & \multirow{2}[2]{*}{2020} & \multicolumn{2}{l}{CMU-MOSI} & \multicolumn{2}{l}{84.10\%} \\
    \multicolumn{2}{l}{} &       & \multicolumn{2}{l}{CMU-MOSEI} & \multicolumn{2}{l}{84.50\%} \\
    \midrule
    \multicolumn{2}{l}{\multirow{2}[2]{*}{TIMF}} & \multirow{2}[2]{*}{2021} & \multicolumn{2}{l}{CMU-MOSI} & \multicolumn{2}{l}{92.28\%} \\
    \multicolumn{2}{l}{} &       & \multicolumn{2}{l}{CMU-MOSEI} & \multicolumn{2}{l}{79.46\%} \\
    \midrule
    \multicolumn{2}{l}{Auto-ML based Fusion} & 2021  & \multicolumn{2}{l}{B-T4SA} & \multicolumn{2}{l}{95.19\%} \\
    \midrule
    \multicolumn{2}{l}{\multirow{3}[2]{*}{Self-MM}} & \multirow{3}[2]{*}{2022} & \multicolumn{2}{l}{CMU-MOSI} & \multicolumn{2}{l}{84.00\%} \\
    \multicolumn{2}{l}{} &       & \multicolumn{2}{l}{CMU-MOSEI} & \multicolumn{2}{l}{82.81\%} \\
    \multicolumn{2}{l}{} &       & \multicolumn{2}{l}{CH-SIMS} & \multicolumn{2}{l}{80.74\%} \\
    \midrule
    \multicolumn{2}{l}{\multirow{2}[2]{*}{DISRFN}} & \multirow{2}[2]{*}{2022} & \multicolumn{2}{l}{CMU-MOSI} & \multicolumn{2}{l}{83.60\%} \\
    \multicolumn{2}{l}{} &       & \multicolumn{2}{l}{CMU-MOSEI} & \multicolumn{2}{l}{87.50\%} \\
    \midrule
    \multicolumn{2}{l}{\multirow{2}[2]{*}{TETFN}} & \multirow{2}[2]{*}{2023} & \multicolumn{2}{l}{CMU-MOSI} & \multicolumn{2}{l}{84.05\%} \\
    \multicolumn{2}{l}{} &       & \multicolumn{2}{l}{CMU-MOSEI} & \multicolumn{2}{l}{84.25\%} \\
    \midrule
    \multicolumn{2}{l}{\multirow{2}[2]{*}{TEDT}} & \multirow{2}[2]{*}{2023} & \multicolumn{2}{l}{CMU-MOSI} & \multicolumn{2}{l}{89.30\%} \\
    \multicolumn{2}{l}{} &       & \multicolumn{2}{l}{CMU-MOSEI} & \multicolumn{2}{l}{86.20\%} \\
    \midrule
    \multicolumn{2}{l}{\multirow{3}[2]{*}{SPIL}} & \multirow{3}[2]{*}{2023} & \multicolumn{2}{l}{CMU-MOSI} & \multicolumn{2}{l}{85.06\%} \\
    \multicolumn{2}{l}{} &       & \multicolumn{2}{l}{CMU-MOSEI} & \multicolumn{2}{l}{85.01\%} \\
    \multicolumn{2}{l}{} &       & \multicolumn{2}{l}{CH-SIMS} & \multicolumn{2}{l}{81.25\%} \\
    \bottomrule
    \end{tabular}%
    \caption{This table contains some of the most recent and top-performing multimodal sentiment analysis models. The first column is the name of the model. The second column is the year in which the model was published. The third column is the dataset used by the model. The fourth column is the accuracy under this dataset.}
  \label{tab:addlabe2}%
\end{table*}%

\subsubsection{\textbf{MultiSentiNet-Att\cite{xu2017multisentinet}}} This model uses an LSTM network to incorporate text information into word vectors. VGG is used to extract both target feature information and scene feature information of an image. The target and scene feature vectors are used to perform cross-modal attention mechanism learning with word vectors. That is, the target feature information and the scene feature information of the image are combined to assign special weights to the word vectors related to the sentiment in the text. The resulting features are fed into a multi-layer perceptron to complete the sentiment analysis task.

\subsubsection{\textbf{DFF-ATMF\cite{chen2019complementary}}} The proposed model mainly considers text modality and audio modality. The main contribution is to propose new multi-feature fusion strategies and multi-modal fusion strategies. Two parallel branches are used to learn features for text modality and features for audio modality. For the features of these two modalities, a multimodal attention fusion module is used to complete the multimodal fusion.

\subsubsection{\textbf{AHRM\cite{xu2020social}}} This model is mainly used to capture the relationship between text modality and visual modality. The authors propose an attention mechanism based heterogeneous relation model, which can well integrate the respective high-quality representation information of text modality and visual modality. This progressive dual attention mechanism can well highlight the channel-level semantic information of image and text information. To integrate social attributes, social relations are introduced to capture complementary information from the social context, and heterogeneous networks are constructed to integrate features.

\subsubsection{\textbf{SFNN\cite{wu2020sfnn}}} The proposed model is a neural network based on semantic feature fusion. Convolutional neural networks and attention mechanisms are used to extract visual features. Visual features are mapped to text features and combined with text modality features for sentiment analysis.

\subsubsection{\textbf{MISA\cite{hazarika2020misa}}} The proposed model presents a novel multi-modal sentiment analysis framework. Each modality is mapped into two distinct feature spaces after feature extraction. One feature space mainly learns the invariant features of the modality and the other one learns the unique features of the modality.

\subsubsection{\textbf{MAG-BERT\cite{rahman2020integrating}}} The authors propose a "multi-modal" adaptation architecture and apply it to BERT. The model can receive input from multiple modalities during fine-tuning. MAG can be thought of as a vector embedding structure that allows us to input multimodal information and embed it as a sequence to BERT.

\subsubsection{\textbf{TIMF\cite{sun2021two}}} The main idea of this model is that each modality learns features separately and performs tensor fusion of the features of each modality. In the dataset fusion stage, the feature fusion for each modality is implemented by a tensor fusion network. In the decision fusion stage, the upstream results are fused by soft fusion to adjust the decision results.

\subsubsection{\textbf{Auto-ML based Fusion\cite{lopes2021automl}}} The authors propose to combine text and image individual sentiment analysis into a final fused classification based on AutoML. This approach combines individual classifiers into a final classification using the best model generated by Auto-ML. This is a typical model for decision-level fusion.

\subsubsection{\textbf{Self-MM\cite{yu2021learning}}} In, the authors combine self-supervised learning and multi-task learning to construct a novel multi-modal sentiment analysis architecture. To learn the private information of each modality, the authors construct a single-modal label generation module ULGM based on self-supervised learning. The loss function corresponding to this module is designed to incorporate the private features learned by the three self-supervised learning subtasks into the original multi-modal sentiment analysis model using a weight adjustment strategy. The proposed model performs well, and the self-supervised learning based ULGM module also has the ability of single-modal label calibration.

\subsubsection{\textbf{DISRFN\cite{he2022dynamic}}} The model is a dynamically invariant representation-specific fusion network. The joint domain separation network is improved to obtain a joint domain separation representation for all modalities, so that redundant information can be effectively utilized. Second, a HGFN network is used to dynamically fuse the feature information of each modality and learn the features of multiple modal interactions. At the same time, a loss function that improves the fusion effect is constructed to help the model learn the representation information of each modality in the subspace.

\subsubsection{\textbf{TEDT\cite{wang2023tedt}}} This model proposes a multimodal encoding-decoding translation network with a transformer to address the challenges of multimodal sentiment analysis, specifically the impact of individual modal data and the poor quality of nonnatural language features. The proposed method uses text as the primary information and sound and image as the secondary information, and a modality reinforcement cross-attention module to convert nonnatural language features into natural language features to improve their quality. Additionally, a dynamic filtering mechanism filters out error information generated in the cross-modal interaction. The strength of this model lies in its ability to improve the effect of multimodal fusion and more accurately analyze human sentiment. However, it may require significant computational resources and may not be suitable for real-time analysis.

\subsubsection{\textbf{TETFN\cite{wang2023tetfn}}} The Text Enhanced Transformer Fusion Network (TETFN) is a novel method proposed for multimodal sentiment analysis (MSA) that addresses the challenge of different contributions of textual, visual, and acoustic modalities.  The proposed method learns text-oriented pairwise cross-modal mappings for obtaining effective unified multimodal representations.  It incorporates textual information in learning sentiment-related nonlinguistic representations through text-based multi-head attention and retains differentiated information among modalities through unimodal label prediction.  Additionally, the vision pre-trained model Vision-Transformer is utilized to extract visual features from the original videos to preserve both global and local information of a human face. The strength of this model lies in its ability to incorporate textual information to improve the effectiveness of nonlinguistic modalities in MSA, while preserving inter- and intra-modality relationships. 

\subsubsection{\textbf{SPIL\cite{lai2023shared}}} This model proposes a deep modal shared information learning module for effective representation learning in multimodal sentiment analysis tasks.   The proposed module captures both shared and private information in a complete modal representation by using a covariance matrix to capture shared information between modalities and a self-supervised learning strategy to capture private information.   The module is plug-and-play and can adjust the information exchange relationship between modes to learn private or shared information.   Additionally, a multi-task learning strategy is employed to help the model focus its attention on modal differentiation training data.   The proposed model outperforms current state-of-the-art methods on most metrics of three public datasets, and more combinatorial techniques for the use of the module are explored.

\section{Model comparison and suggestions}

This section evaluates five state-of-the-art multimodal sentiment analysis models: DFF-ATMF, MAG-BERT, TIMF, Self-MM, and DISRFN. While DFF-ATMF does not consider the vision modality, the other models analyze sentiment from three modalities of audio, text, and vision.

For the interaction relations of multimodal data, DFF-ATMF and TIMF build transformer-based models to learn complex relationships among the data. MAG-BERT uses a simple yet effective multimodal adaptive gate fusion strategy. Self-MM uses self-supervised multi-task learning as the fusion strategy, self-supervised generation of single-modal labels, and combination of single-modal labels to complete the multi-modal sentiment analysis task. DISRFN uses a Dynamic Invariant-Specific Representation Fusion Network to obtain jointly domain-separated representations of all modalities and dynamically fuse each representation through a hierarchical graph fusion network.

\textbf{DFF-ATMF} uses two parallel branches to fuse audio and text modalities. Its core mechanisms are feature vector fusion and multimodal attention fusion, which can learn more comprehensive sentiment information. However, due to the use of multi-layer neural networks and sophisticated fusion methods, overfitting may occur.
Advantages: Simple structure, easy to implement, and can learn comprehensive sentiment information through feature vector fusion and multimodal attention fusion.
Disadvantages: Does not consider the vision modality, may suffer from overfitting due to the use of multi-layer neural networks and sophisticated fusion methods.

\textbf{MAG-BERT} adapts the interior of BERTs using multimodal adaptation gates, which employ a simple yet effective fusion strategy without changing the structure and parameters of BERTs. However, the multimodal attention can only be performed within the same timestep but not across timesteps, which may ignore some temporal relationships. Additionally, MAG-BERT requires freezing the parameters of BERT without being able to fine-tune BERT, which may result in a representation of BERT that is not adapted to a specific task or domain.
Advantages: Uses a simple yet effective multimodal adaptive gate fusion strategy without changing the structure and parameters of BERTs.
Disadvantages: Multimodal attention can only be performed within the same timestep but not across timesteps, which may ignore some temporal relationships. Requires freezing the parameters of BERT without being able to fine-tune BERT, which may result in a representation of BERT that is not adapted to a specific task or domain.

\textbf{TIMF} leverages the self-attention mechanism of Transformers to learn complex interactions between multimodal data and generate unified sentiment representations. While it has the advantage of being able to learn complex relationships between modalities, it may suffer from extreme computational complexity, long training times, and problems with large amounts of labeled data.
Advantages: Leverages the self-attention mechanism of Transformers to learn complex interactions between multimodal data and generate unified sentiment representations.
Disadvantages: May suffer from extreme computational complexity, long training times, and problems with large amounts of labeled data.

% Table generated by Excel2LaTeX from sheet 'table1'
\begin{table*}[htbp]
  \centering
    \begin{tabular}{ccccccccc}
    \toprule
    \multirow{2}[4]{*}{\textbf{Model}} & \multicolumn{4}{c}{\textbf{CMU-MOSI}} & \multicolumn{4}{c}{\textbf{CMU-MOSEI}} \\
\cmidrule{2-9}          & \textbf{MAE} & \textbf{Corr} & \textbf{Acc} & \textbf{F1-Score} & \textbf{MAE} & \textbf{Corr} & \textbf{Acc} & \textbf{F1-Score} \\
    \midrule
    DFF-ATMF & --    & --    & 80.9  & 81.3  & --    & --    & 77.2  & 78.3 \\
    MAG-BERT & 0.712 & 0.796 & --    & 86    & 0.623 & 0.677 & 82    & 82.1 \\
    TIMF  & \textbf{0.373} & \textbf{0.93} & \textbf{92.3} & \textbf{92.3} & 0.645 & 0.669 & 79.5  & 79.5 \\
    Self-MM & 0.723 & 0.797 & 84.8  & 84.8  & 0.534 & 0.764 & 84.1  & 84.1 \\
    DISRFN & 0.798 & 0.734 & 83.4  & 83.6  & 0.591 & \textbf{0.78} & \textbf{87.5} & \textbf{87.5} \\
    TEDT & 0.709 & 0,812 & 0.893 & 0.892 & 0.524 & 0.749 & 0.862 & 0.861 \\
    TETFN & 0.717 & 0.800 & 0.841 & 0.838 & 0.551 & 0.748 & 0.843 & 0.842 \\
    SPIL  & 0.704 & 0.794 & 0.851 & 0.854 & \textbf{.523} & 0.766 & 0.850 & 0.849  \\
    
    \bottomrule
    \end{tabular}%
     \caption{The table shows the performance metrics of the DFF-ATMF, MAG-BERT, TIMF, Self-MM, DISRFN, TEDT ,TETFN and SPIL models under the CMU-MOSI and CMU-MOSEI datasets. The evaluation parameters included: MAE, Corr, Acc and F1-Score.}
  \label{tab:addlabe3}%
\end{table*}%

\textbf{Self-MM} is a self-supervised multi-modal sentiment analysis model that uses a multi-task learning strategy to learn both multimodal and unimodal emotion recognition tasks. Its advantage is that it can generate single-modal labels using a self-supervised approach, saving the cost and time of manual labeling. However, interference and imbalance between multiple tasks can occur, and an appropriate weight adjustment strategy needs to be designed to balance the learning progress of different tasks.
Advantages: Self-supervised multi-task learning as the fusion strategy, self-supervised generation of single-modal labels, and combination of single-modal labels to complete the multi-modal sentiment analysis task.
Disadvantages: May require a large amount of labeled data to achieve good performance, and the self-supervised learning process may be computationally expensive.

\textbf{DISRFN} is a deep residual network-based multi-modal sentiment analysis model that exploits the strategy of Dynamic Invariant-Specific Representation Fusion Network to improve sentiment recognition capability. Its advantage is that it can efficiently utilize redundant information to obtain joint domain-separated representations of all modalities through a modified joint domain separation network and dynamically fuse each representation through a hierarchical graph fusion network to obtain the interaction information of multimodal data. However, as with Self-MM, interference and imbalance between multiple tasks can occur, and a suitable weight adjustment strategy needs to be designed to balance the learning progress of different tasks.
Advantages: Uses a Dynamic Invariant-Specific Representation Fusion Network to obtain jointly domain-separated representations of all modalities and dynamically fuse each representation through a hierarchical graph fusion network.
Disadvantages: May require a large amount of labeled data to achieve good performance, and the hierarchical graph fusion network may be computationally expensive.

\textbf{TEDT} proposes a multimodal encoding-decoding translation network with a transformer to address the challenges of multimodal sentiment analysis. The strength of this model lies in its ability to improve the effect of multimodal fusion and more accurately analyze human sentiment.  By incorporating the modality reinforcement cross-attention module and dynamic filtering mechanism, the model is able to address the challenges of individual modal data impact and poor quality of nonnatural language features. To build effective multimodal sentiment analysis models, it is recommended to carefully consider the contribution of each modality and how to effectively integrate them.  Additionally, attention should be paid to addressing challenges such as individual modal data impact and poor quality of nonnatural language features.  Finally, it is important to consider the computational requirements of the model and ensure that it is suitable for the intended use case.

\textbf{TETFN} is a novel method proposed for MSA that addresses the challenge of different contributions of textual, visual, and acoustic modalities.   Compared to the TEDT model, the TETFN model focuses on incorporating textual information to improve the effectiveness of nonlinguistic modalities in MSA while preserving inter- and intra-modality relationships.    The TETFN model achieves this by using text-based multi-head attention and unimodal label prediction to retain differentiated information among modalities.    In contrast, the TEDT model uses a modality reinforcement cross-attention module to convert nonnatural language features into natural language features and a dynamic filtering mechanism to filter out error information generated in the cross-modal interaction.  The strength of the TETFN model lies in its ability to effectively incorporate textual information to improve the effectiveness of nonlinguistic modalities in MSA while preserving inter- and intra-modality relationships.    Additionally, the use of the vision pre-trained model Vision-Transformer helps to extract visual features from the original videos to preserve both global and local information of a human face. To build effective multimodal sentiment analysis models, it is recommended to carefully consider the contribution of each modality and how to effectively integrate them. Additionally, attention should be paid to addressing challenges such as individual modal data impact and poor quality of nonnatural language features.

\textbf{SPIL} proposes a deep modal shared information learning module for effective representation learning in multimodal sentiment analysis tasks.   The proposed module captures both shared and private information in a complete modal representation by using a covariance matrix to capture shared information between modalities and a self-supervised learning strategy to capture private information.   The module is plug-and-play and can adjust the information exchange relationship between modes to learn private or shared information.   Additionally, a multi-task learning strategy is employed to help the model focus its attention on modal differentiation training data.  Compared to the TEDT and TETFN models, the SPIL model also focuses on capturing shared and private information in a complete modal representation.   However, the SPIL model uses a covariance matrix to capture shared information between modalities and a self-supervised learning strategy to capture private information, while the TETFN model uses text-based multi-head attention and unimodal label prediction to retain differentiated information among modalities.   The SPIL model also employs a multi-task learning strategy to help the model focus its attention on modal differentiation training data, while the TEDT and TETFN models do not explicitly mention this. The strength of the SPIL model lies in its ability to capture both shared and private information in a complete modal representation, which can be adjusted based on the specific task at hand.   Additionally, the use of a multi-task learning strategy helps to improve the model's performance by focusing its attention on modal differentiation training data.  The SPIL model's approach of capturing both shared and private information in a complete modal representation is worth considering in future models.

Tab\ref{tab:addlabe3} shows the performance metrics of these five models under the CMU-MOSI and CMU-MOSEI datasets. When analyzing the performance metrics of the five models on these datasets, we recommend using BERT to extract features of text information while using LSTM to extract features for video and audio modality information since it requires capturing modality information in the time series.

DFF-ATMF does not consider visual modality, resulting in relatively low performance metrics. Visual information can provide additional information about human expressions, poses, scenes, etc., which can enhance the information of text and speech modalities as well as complement them. Therefore, visual modality information deserves to be considered and explored in multimodal sentiment analysis.

\section{Challenges and Future Scope}
With the development of deep learning, multimodal sentiment analysis techniques have also been rapidly developed\cite{huddar2019survey,kaur2022multimodal,stappen2021multimodal,illendula2019multimodal}. However, multi-modal sentiment analysis still faces many challenges. This subsection analyzes the current state of research, challenges, and future developments in multimodal sentiment analysis.

\subsection{Dataset}
In multimodal sentiment analysis, the dataset plays a crucial role. Currently, a large dataset in multiple languages is missing. Given the diversity of languages and races in many countries, a large, diverse dataset could be used to train a multi-modal sentiment analysis model with strong generalization and wide usage. Additionally, current multimodal datasets still have low annotation accuracy and have not yet reached absolute continuous values, requiring researchers to label multimodal datasets more finely. Most current multimodal data contain only visual, speech, and text modalities and lack modal information combined with physiological signals such as brain waves and pulses.

\subsection{Detection of Hidden Emotions}
There has always been a recognized difficulty in multimodal sentiment analysis tasks: the analysis of hidden emotions. Hidden emotions\cite{tang2019hidden,hajek2020fake} include: sarcastic emotions (such as sarcastic words), emotions that need to be concretely analyzed in context, and complex emotions\cite{kwon2019cnn,rashid2020emotion} (such as a person's happiness and sadness). It is important to explore these hidden emotions. It's the gap between human and artificial intelligence\cite{ghanbari2019text}.

\subsection{Multiple forms of video data}
In multimodal sentiment analysis tasks, video data is particularly challenging. Although the speaker is facing the camera and the video data resolution is maintained at a high level, the actual situation is more complicated and requires the model to be robust against noise and applicable to low-resolution video data. Capturing the micro-expressions and micro-gestures of speakers for sentiment analysis is also an area worth exploring by researchers.

\subsection{Multiform language data}
The form of text data in multimodal sentiment analysis tasks is typically single. However, evaluation texts in online communities are often cross-lingual, with reviewers using multiple languages to make more vivid comments. Text data with mixed emotions also remains a challenge for multimodal sentiment analysis tasks. Making good use of memes that are mixed in the text is an important research topic, as memes often contain extremely strong emotional messages about reviewers. Additionally, most text data is transcribed directly through speech, making it particularly difficult to analyze a person's emotions when multiple people are talking. Combined with the cultural characteristics of different regions and countries, the same text data may reflect different emotions.

\subsection{Future Prospectsn}
The future of multimodal sentiment analysis techniques is extremely bright, and some of the future applications are listed below. Multimodal emotion analysis for real-time assessment of mental health\cite{xu2020inferring,walambe2021employing,aloshban2022you};  multimodal criminal linguistic deception detection model\cite{chebbi2021deception};  offensive language detection;  A human-like emotion-aware robot, etc. Multimodal emotion analysis is a technique for recognizing and analyzing emotions. Models that combine multi-modal information data for sentiment analysis can effectively improve the accuracy of sentiment analysis. In the future, multi-modal sentiment analysis techniques will be gradually improved. Perhaps one day there will be a multimodal sentiment analysis model with a large number of parameters that will have the same sentiment analysis capabilities as humans. It was a thing of rapture.

\section{Conclusion}
Multimodal sentiment analysis techniques have been recognized as important by researchers in various fields, making it a central research topic in the fields of natural language processing and computer vision. In this review, we provide a detailed description of various aspects of multimodal sentiment analysis, including its research background, definition, and development process. We also summarize commonly used benchmark datasets in Table 1 and compare and analyze recent state-of-the-art multimodal sentiment analysis models. Finally, we present the challenges posed by the field of multimodal sentiment analysis and explore possible future developments.

Many prospective works are being actively carried out and have even been largely implemented. However, there are still challenges to be addressed, leading to the following meaningful research directions:

(1) Construct a large multimodal sentiment dataset in multiple languages.

(2) Solve the domain transfer problem of video, text, and speech modal data.

(3) Build a unified, large-scale multimodal sentiment analysis model with excellent generalization performance.

(4) Reduce model parameters, optimize algorithms, and reduce algorithm complexity.

(5) Solve the multi-lingual hybridness problem in multimodal sentiment analysis.

(6) Discuss the weight problem of modal fusion and provide the most reasonable scheme to assign weights of different modalities in different cases.

(7) Discuss the correlation between modalities and separate shared and private information between them to improve model performance and interpretability.

(8) Construct a multimodal sentiment analysis model that can well complete hidden emotions.

\section*{Declarations}
\subsection*{Availability of data and materials}
Not applicable.
\subsection*{Competing interests}
The authors declare that they have no competing interests.
\subsection*{Funding}
This work was supported in part by Joint found for smart computing of Shandong Natural Science Foundation under Grant ZR2020LZH013; open project of State Key Laboratory of Computer Architecture CARCHA202001; the Major Scientific and Technological Innovation Project in Shandong Province under Grant 2021CXG010506 and 2022CXG010504; "New University 20 items" Funding Project of Jinan under Grant 2021GXRC108 and 2021GXRC024.
% \subsection*{Authors' contributions}
% Songning Lai completed the production of pictures, tables and the writing of the whole paper. Haoxuan Xu, Xifeng Hu, Zhaoxia Ren and Zhi Liu reviewed drafts and edited the fnal copy.  All authors read and approved the  fnal manuscript.
\subsection*{Acknowledgments}
Not applicable.
% BibTeX from reference.bib
%\bibliographystyle{sn-apacite}
\bibliographystyle{unsrt}
\bibliography{reference}

\end{document}